\documentclass{article}

\usepackage{microtype}
\usepackage{graphicx}
\usepackage[percent]{overpic}
\usepackage{hyperref}
\usepackage{stfloats}
\usepackage[accepted]{icml2026}

\usepackage{amsmath,amsfonts,bm}

\def\eqref#1{equation~\ref{#1}}

\def\1{\bm{1}}

\DeclareMathAlphabet{\mathsfit}{\encodingdefault}{\sfdefault}{m}{sl}
\SetMathAlphabet{\mathsfit}{bold}{\encodingdefault}{\sfdefault}{bx}{n}

\usepackage{amsmath}
\usepackage{amssymb}
\usepackage{mathtools}
\usepackage[capitalize,noabbrev]{cleveref}

\icmltitlerunning{Device Passport for Cross-Layout Biosignal Transfer}

\begin{document}

\twocolumn[
  \icmltitle{Device Passport: Enabling Spatio-Temporal Pretrained Models to Generalize Across Input Layouts}

  \begin{icmlauthorlist}
    \icmlauthor{Geeling Chau}{caltech,apple}
    \icmlauthor{Ran Liu}{apple}
    \icmlauthor{Juri Minxha}{apple}
    \icmlauthor{Wenhui Cui}{apple}
    \icmlauthor{Erdrin Azemi}{apple}
    \icmlauthor{Ellen L. Zippi}{apple}
    \icmlauthor{Behrooz Mahasseni}{apple}
    \icmlauthor{Christopher Michael Sandino}{apple}
  \end{icmlauthorlist}

  \icmlaffiliation{caltech}{California Institute of Technology, Pasadena, CA, USA}
  \icmlaffiliation{apple}{Apple, Cupertino, CA, USA}
  \icmlcorrespondingauthor{Christopher Michael Sandino}{csandino@apple.com}

  \icmlkeywords{biosignal foundation models, EEG, layout transfer, channel embeddings}

  \vskip 0.3in
]

\printAffiliationsAndNotice{}

\begin{abstract}
New device layouts pose a challenging modeling problem due to the lack of large datasets for each specific layout. 
Biosignal foundation models offer a plausible solution if they are able to generalize to new layouts effectively. 
To improve cross-layout transfer, we study how different channel embedding techniques behave when pretraining layouts differ substantially from the downstream decoding layout.
We propose Device Passport, a new channel embedding technique that learns experts and mixture models that take each channel's functional activity and metadata as input.
This contrasts with prior embedding methods, which typically use only functional information or only metadata to look up learned or fixed positional embeddings.
Across controlled subset-transfer experiments and realistic transfer to ear-EEG, Device Passport is competitive overall and improves over the strongest learned baseline in the layout-transfer regimes that motivate this work.
These results suggest that channel embedding design is a key consideration when reusing large-scale pretrained biosignal models on new devices.
\end{abstract}

\section{Introduction}
Across biosignal domains such as EMG, EKG, and EEG, large-scale biosignal foundation models have been shown to improve downstream decoding, especially when pretraining and evaluation share similar sensor layouts \citep{kaifosh2025generic,abbaspourazad2023large, wang2024cbramod}.
However, it is common for channels to be placed in new locations when experimenting with new devices, constituting unseen sensor layouts. 
In these settings, extensive pretraining data may not be available or only be available in mismatched layouts, so it would be valuable for such pretrained models to generalize to new device layouts. 
This problem is central to wearable health sensing, where new form factors often collect high-frequency physiological time series before large device-specific cohorts exist.
For many prototype or emerging health devices, collecting large labeled cohorts for every new montage is impractical, so the useful question is not only whether a pretrained biosignal model works, but whether its spatial knowledge can be reused when the sensor layout changes.

A core challenge in this setting is channel embedding.
Transformers offer channel count flexibility, but rely on positional or channel embeddings to contextualize each input \cite{vaswani2017attention, chau2025population}. 
Most positional or lookup-based embedding schemes assume repeated identities or enough examples per channel to relearn useful embeddings after transfer.
Under strong layout shift and low-data experimental settings, these assumptions break, and the model must recover spatial relationships for sensor channels whose locations or identities were not available during pretraining.
We therefore argue that channel embeddings are a core weakness in unseen-layout transfer.

Prior channel embedding methods span identity-based, coordinate-based, and activity-conditioned schemes, including strong recent approaches such as asymmetric channel positional embedding (ACPE) \cite{wang2024cbramod}.
However, it remains unclear how well these methods and ACPE transfer when the downstream montage differs or data are scarce.
This motivates a focused study of how to best learn and initialize channel embeddings when layout transfer is large.
In particular, we study whether channel embeddings should be derived only from functional activity, or whether they should instead be estimated from both metadata and functional activity.

In this work, we systematically compare six channel embedding strategies, including sinusoidal, channel ID, XYZ-based encodings, ACPE, and two novel expert-based adapters in the Device Passport family that learn to leverage metadata and functional activity of new channels to estimate channel embeddings.
We evaluate these methods in two settings: a controlled layout-transfer experiment based on pretraining and fine-tuning on different TUH/TUAB electrode subsets, and a realistic transfer problem from full-layout TUH pretraining to ear-EEG sleep staging on EESM17.
Across both settings, Device Passport is most useful when transferring to new layouts, suggesting a practical path for reusing spatial knowledge across devices without relearning channel embeddings from scratch.

\section{Approach and Methods}
\textbf{Models.}
We conduct our experiments on a single common backbone model, the CBraMod EEG model \cite{wang2024cbramod}, which is a spatiotemporal transformer pretrained using masked reconstruction of spatiotemporal patches.
A key component of CBraMod is the channel embedding technique, Asymmetric Conditional Positional Encoding (ACPE), which dynamically encodes spatial and temporal position.
In our experiments, we keep the original pretraining pipeline and preprocessing (e.g., filtering and data scaling), and vary only the channel embedding technique across a broader family of alternatives, including ACPE and simpler lookup-based encodings.

\begin{figure}[b!]
  \centering
  \includegraphics[width=\linewidth]{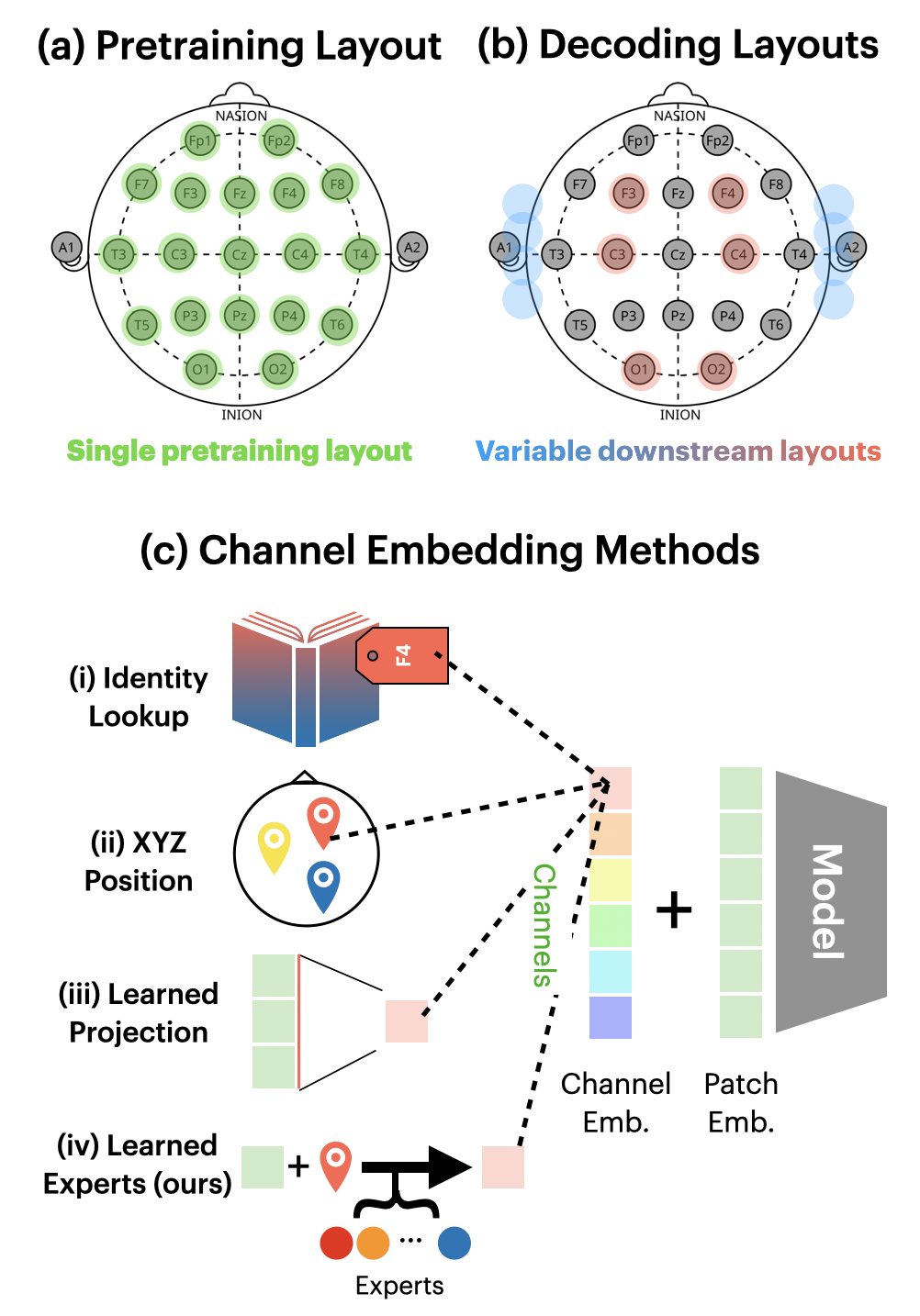}
  \caption{
  \textbf{Layout Transfer Challenge + Channel Embedding Techniques.}
  (a) Pretraining often occurs on a single pretraining layout.
  (b) Decoding needs to work on variable downstream layouts.
  (c) Channel embedding methods help identify the origin of functional activity, but many techniques do not learn transferable representations due to channel layout mismatch between pretraining and decoding.
  (c.i) Identity lookup channel embeddings can be learned (channel id) or fixed (APE).
  (c.ii) XYZ position-based lookup as used in \cite{chau2025population}.
  (c.iii) Learned projection (ACPE) as used in \cite{wang2024cbramod}.
  (c.iv) Our method learns mixture models and expert embeddings during pretraining, which can be leveraged during downstream decoding.
  Device Passport denotes the learned expert-adapter variants shown in (c.iv).
  \label{position_encoding_figure}
  }
\end{figure}

\begin{figure*}[!t]
  \centering
  \includegraphics[width=0.95\textwidth]{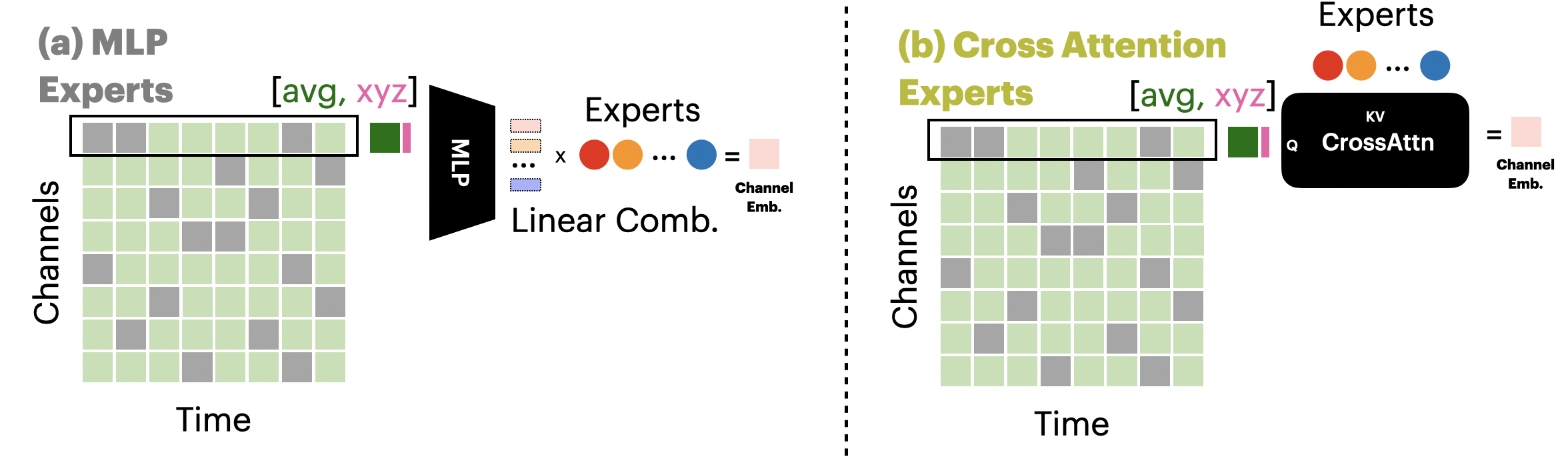}
  \caption{
  \textbf{Our method.}
We learn (a) MLP or (b) Cross-Attention mechanisms that combine channel activity with XYZ location to weight or attend to a set of learned experts.
  \label{method_figure}
  }
\end{figure*}

\textbf{Channel Embedding Techniques.}
We investigate six different positional encoding techniques.
Four are baselines (APE, Channel ID, XYZ, ACPE) and two are variations of our proposed expert-based channel embedding adapters (MLP Experts and Cross-Attention Experts).
We use Device Passport to refer to this expert-adapter family, where MLP and Cross-Attention variants describe the mixture model architecture used.
The baselines span ordinal (APE), identity (Channel ID), coordinate-based (XYZ), and activity-conditioned (ACPE) encodings \citep{vaswani2017attention, azabou2023unified, jiang2024large, wang2024cbramod}.
Device Passport uses channel location (XYZ) and functional patch embeddings as inputs to a mixture model for producing a channel embedding.
Concretely, for channel \(c\), we pool non-masked patch embeddings into an activity summary \(z_c\), concatenate it with metadata \(m_c\), and compute a channel embedding from learned experts \(E_k\).
In the MLP variant, we compute channel-specific mixture weights over experts, \(w_c=\operatorname{softmax}(\operatorname{MLP}([z_c,m_c]))\), and use them to form \(e_c=\sum_k w_{c,k}E_k\), where \(w_{c,k}\) denotes the weight assigned to expert \(k\). The cross-attention variant instead uses \([z_c,m_c]\) as a query over learned expert keys/values.
More details of each are available in \Cref{position_encoding_figure,method_figure} and Appendix~\ref{app:channel-embeddings}.

By deriving positional encodings from pretrained expert embeddings and mixture models, Device Passport uses pretrained expert embeddings to contextualize new electrode layouts.
During pretraining, Device Passport learns a bank of 10 expert embeddings (\(E_k; k=1,\ldots,10\)) together with either an MLP or a Cross-Attention mixture module.
At downstream transfer time, the expert embeddings are frozen, while the mixture module is fine-tuned so that the model can reuse pretrained expert embeddings while adapting the mixture rule to the new layout.
This differs from ACPE, which derives positional embeddings from functional activity and not additional channel metadata.
Our goal is to test whether explicitly using pretrained expert embeddings to contextualize new electrode layouts could benefit downstream decoding performance, especially for electrode layouts unseen during pretraining.

\textbf{Pretraining.} \label{pretraining_methods}
We follow CBraMod's pipeline with hyperparameters and preprocessing of data \cite{wang2024cbramod} closely.
The dataset used is the Temple University Hospital (TUH) EEG Corpus \cite{obeid2016temple}, which primarily uses a 19-channel 10--20 electrode configuration across 14k subjects for a total of 27k hours of recording.
Similar to CBraMod's proposed pretraining, we perform data cleaning with a 100~\(\mu\)V cutoff to remove approximately two-thirds of the full dataset, leaving around 9k hours of pretraining data.
This was shown to improve model downstream performance in the original work, so we adopt it for our exploration of different channel embedding techniques.
Across variants, we keep the backbone, objective, and training setup fixed so that differences reflect the channel embedding design.

The main pretraining loss is a reconstruction loss on masked patches.
For adapters (MLP Experts and Cross-Attention Experts) that require patch embedding as input, we use an average of the non-masked patches for each channel.
For ACPE, the patch embeddings across time and space (without distinguishing masked from non-masked patches) are multiplied by the learned CNN kernel to provide the positional embedding.

\textbf{Downstream Decoding.}
We fine-tune using the hyperparameters reported in CBraMod \cite{wang2024cbramod}.
For sample-efficiency sweeps, we match the full fine-tuning compute budget by fixing the number of optimization steps based on CBraMod's epoch schedule.
We evaluate Temple University Abnormal EEG Detection (TUAB) \cite{obeid2016temple}, an abnormal EEG classification task that closely matches the pretraining data family, and EESM17 \cite{mikkelsen2017automatic}, an ear-EEG sleep staging task with a disjoint downstream layout.
For TUAB, the main metric shown in the controlled layout-transfer setting is test AUROC as a function of the number of fine-tuning subjects.
For EESM17, the main figure shows relative Cohen's kappa versus ACPE per held-out subject together with absolute performance across all channel embedding techniques.
We use subject-disjoint train/val/test splits for TUAB and leave-one-subject-out (LOSO) evaluation for EESM17 as proposed in the original work \cite{mikkelsen2017automatic}.
Because ear-EEG channels exhibit different scale and variability, we apply instance normalization to EESM17 signals; otherwise we follow the CBraMod training setup.
Additional model and optimization settings are summarized in Appendix~\ref{app:implementation-details} and Table~\ref{tab:implementation-details}.

\begin{figure}[b!]
  \centering
  \includegraphics[width=\linewidth]{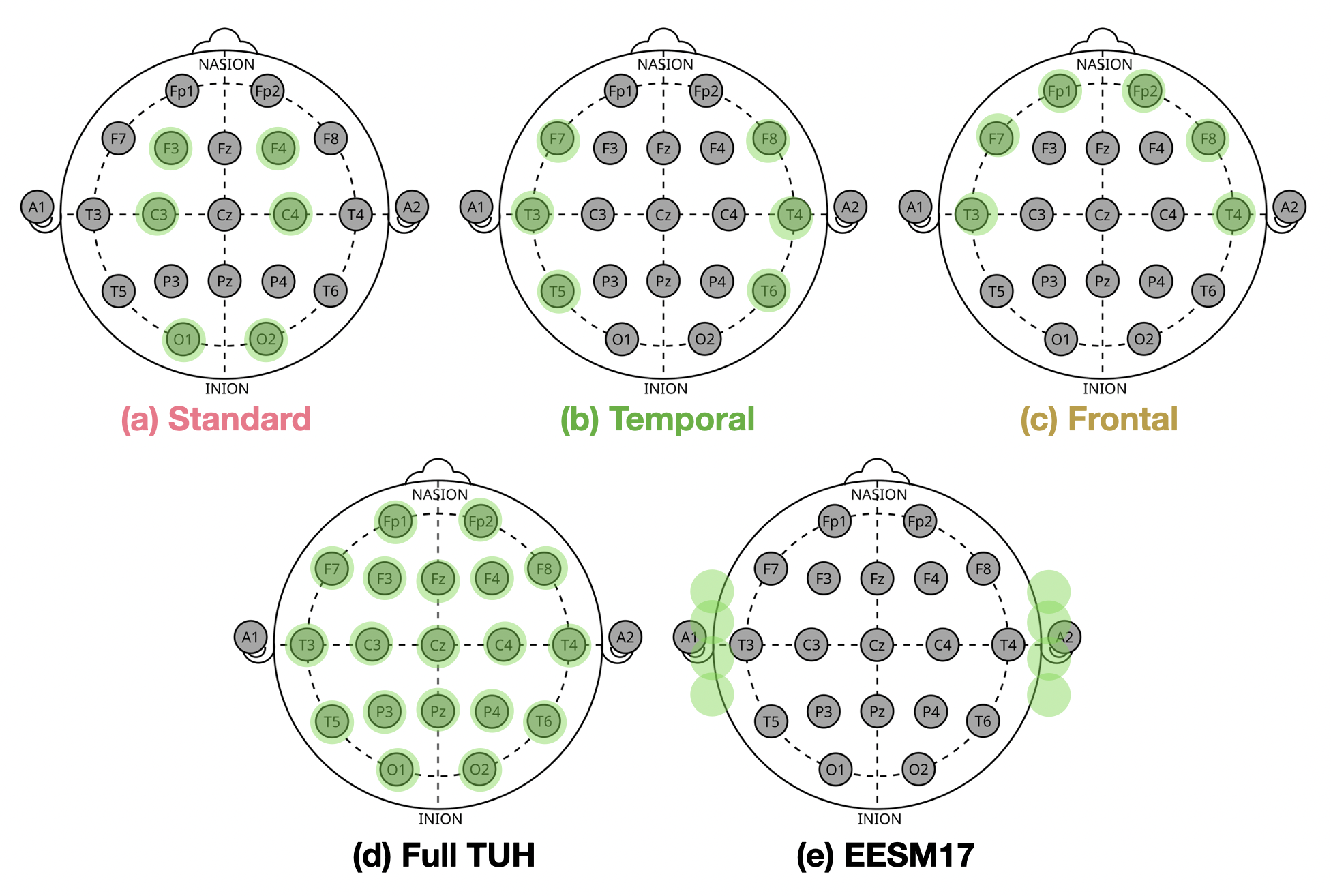}
  \caption{
  \textbf{Electrode layouts.}
  Electrode selections (green) used in our layout transfer experiments, including the TUH subset pretraining layouts (a-c), full pretraining (d), and downstream decoding ear-EEG (e) layouts.
  \label{decoding_layouts_figure}
  }
\end{figure}

\section{Experiments}

\textbf{Variable pretraining layouts and sample efficiency.}
This experiment is a controlled layout-transfer setting: downstream evaluation uses the same Standard TUH/TUAB electrode subset, while the pretraining layout is deliberately varied across Standard, Frontal, and Temporal subsets available from the TUH dataset.
First, we evaluate how existing positional encoding methods degrade in downstream performance with larger layout transfer challenges.
Second, we characterize how different positional embedding techniques perform across fine-tuning sample sizes under different layout transfer challenges.
Each controlled layout subset contains 6 electrodes for consistency in data quantity for the model.
Precise electrode selections are shown in \Cref{decoding_layouts_figure}.
For each layout subset, we train 5 models from different random seeds to cover variability due to random initialization.
Then, we evaluate each of these models on 5 random seeds of fine-tuning on the downstream dataset TUAB \cite{obeid2016temple}.
The downstream dataset uses the Standard layout subset of electrodes, so one pretraining layout (Standard) perfectly matches the downstream dataset, while the other two (Frontal and Temporal) do not.
We focus our sample efficiency sweep on using $<$1/10th ($<$200 out of 2k subjects) of the full TUAB training set.
This setting tests layout transfer in a controlled, data-constrained manner within the same dataset family between pretraining and downstream decoding.

\textbf{Full layout pretraining.}
This experiment is a real-world transfer setting: models are pretrained on the full TUH scalp montage then adapted to a disjoint downstream evaluation dataset, specifically EESM17, an ear-EEG dataset.
The question is whether we see improved transfer to new devices even in this challenging real-world setting where downstream electrode layouts may be completely disjoint from the pretraining set.
Similar to the controlled setting, we pretrain on the full TUH electrode set with five different random seeds.
These models are then evaluated with five random seeds on the downstream dataset EESM17 \cite{mikkelsen2017automatic}.
Because EESM17 uses ear electrodes absent from TUH, this benchmark tests whether pretrained spatial structure can extrapolate beyond the scalp montage.
This provides one of the most challenging layout-transfer problems that practitioners may face when trying to use pretrained models in the real world, and serves as our main real-device transfer benchmark.

\begin{figure*}[!t]
  \centering
  \includegraphics[width=\textwidth]{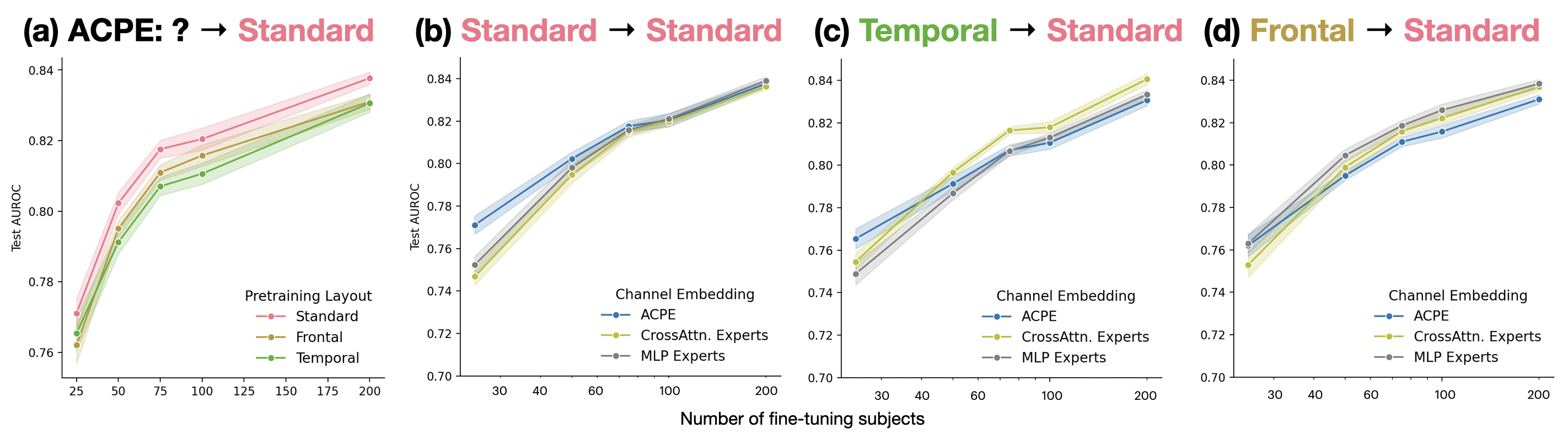}
  \caption{
  \textbf{Variable pretraining layouts.} Downstream decoding performance (AUROC, y-axis) versus number of fine-tuning subjects (x-axis), comparing ACPE across pretraining layouts (a) and channel embedding techniques across matched and mismatched layout-transfer settings (b--d).
  Higher AUROC is better; shaded bands summarize variation across random seeds.
  (a) Pretraining with different layouts (colors) hurts ACPE's ability to transfer performance to the downstream Standard layout.
  (b) Pretrain on Standard, fine-tune on Standard.
  (c) Pretrain on Temporal, fine-tune on Standard.
  (d) Pretrain on Frontal, fine-tune on Standard.
  \label{variable_pretraining_layout_figure}
  }
\end{figure*}

\section{Results}

\begin{figure*}[t]
  \centering
  \begin{overpic}[width=0.81\textwidth]{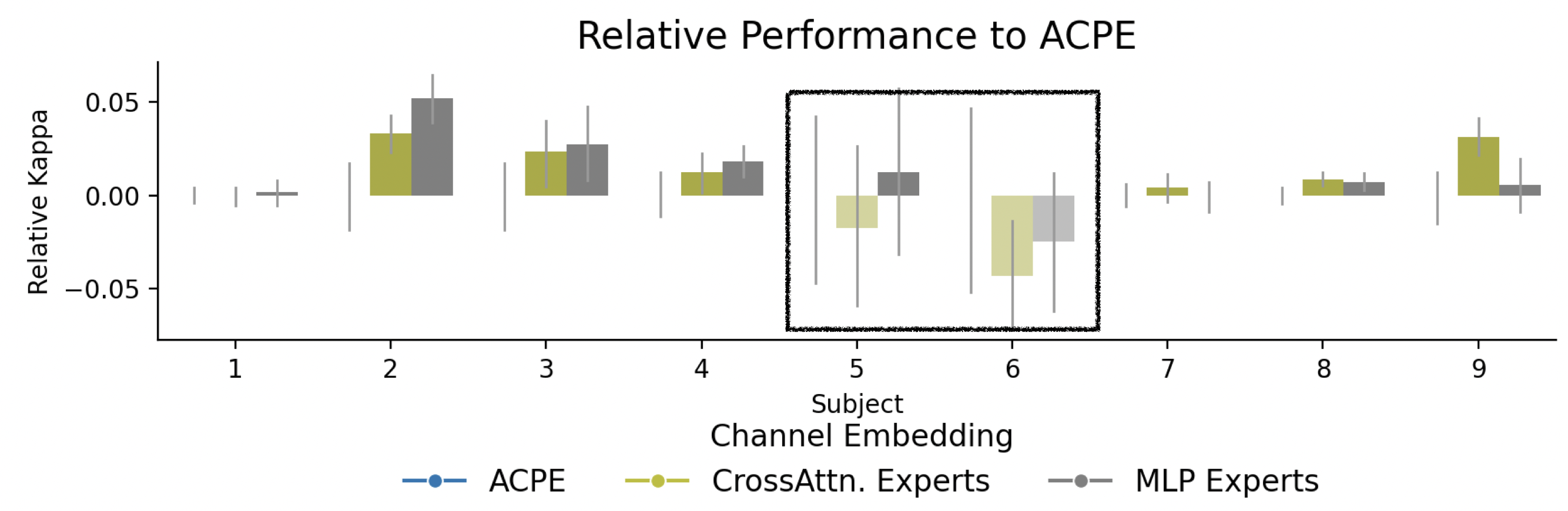}
    \put(1,30){\small\textbf{(a)}}
  \end{overpic}
  \par\vspace{0.6em}
  \begin{minipage}{0.81\textwidth}
    {\small\textbf{(b)}\par}
    \centering
    \resizebox{0.90\linewidth}{!}{%
      \begin{tabular}{l|ccccccccc}
        Model Name & Sub 1 & Sub 2 & Sub 3 & Sub 4 & Sub 5 & Sub 6 & Sub 7 & Sub 8 & Sub 9 \\
        \hline
        ACPE & \underline{0.742} & 0.712 & 0.634 & 0.682 & \underline{0.325} & \textbf{0.356} & 0.579 & 0.791 & 0.666 \\
        CrossAttn. Experts & \underline{0.742} & \underline{0.746} & \underline{0.657} & \underline{0.694} & 0.308 & 0.313 & \textbf{0.583} & \textbf{0.800} & \textbf{0.697} \\
        MLP Experts & \textbf{0.743} & \textbf{0.764} & \textbf{0.661} & \textbf{0.700} & \textbf{0.338} & \underline{0.331} & \underline{0.579} & \underline{0.798} & \underline{0.671} \\
        \hline
      \end{tabular}%
    }
  \end{minipage}
  \caption{
  \textbf{Pretraining on full TUH, fine-tuning on an ear-EEG form factor for sleep staging classification.}
  (a) We plot the performance difference between ACPE and Device Passport across subjects at \(N=6\) training subjects as a focused comparison against the strongest prior learned baseline.
  Positive values indicate that the Device Passport variants improve over ACPE; negative values indicate that ACPE performs better.
  We find that 7/9 subjects in leave-one-subject-out evaluations (x-axis) benefit from Device Passport (positive colored bars).
  The subjects that performed worse with Device Passport (box outline) were previously reported to have poor electrode contact \cite{mikkelsen2017automatic} (subject note in Appendix~\ref{app:eesm17-subject-mapping}).
  Bars show the mean performance difference of the labeled method relative to ACPE, with standard deviation across \(5\) pretraining \(\times\) \(5\) fine-tuning seeds.
  (b) Absolute Cohen's kappa on EESM17 by held-out subject; bold indicates best and underline indicates second best.
  \label{eesm17_subject_figure}
  }
\end{figure*}

\textbf{Variable pretraining layouts and sample efficiency.}
In the controlled TUAB transfer experiment, fine-tuning and held-out testing use the same Standard downstream layout; differences in performance therefore isolate the effect of matched versus mismatched pretraining layouts.
We first confirm that layout mismatch reduces performance: pretraining on Frontal or Temporal subsets and evaluating on Standard is worse than Standard-to-Standard transfer (\Cref{variable_pretraining_layout_figure}a).
We also confirm that ACPE performs on par with or better than traditional channel embedding baselines (Appendix~\ref{app:acpe-vs-traditional}), consistent with prior work \cite{wang2024cbramod}.

We next find that Device Passport adapters often improve over ACPE under layout transfer (\Cref{variable_pretraining_layout_figure}b--d), indicating that improvements can be made to pretraining and adaptation techniques to improve decoding on new device layouts.
In the controlled layout-transfer experiment, Device Passport is most favorable in the mismatched-layout settings that motivate the paper, namely Temporal-to-Standard and Frontal-to-Standard transfer (\Cref{variable_pretraining_layout_figure}c,d).
As the number of fine-tuning subjects increases, performance improves across methods; ACPE tends to be quite strong at the very lowest-data regime, while Device Passport is stronger across most intermediate regimes, especially on layout transfer settings.
We hypothesize that ACPE may be easier to adapt in the smallest fine-tuning regimes because its activity-conditioned convolutional positional update has relatively few trainable parameters compared with our expert-adapter variants, and it avoids explicit metadata-to-expert assignment choices.
Device Passport becomes more useful once there is enough downstream data to tune how activity and metadata select among pretrained expert anchors.

\textbf{Full layout pretraining.}
In the EESM17 transfer experiment, all methods start from full-layout TUH pretraining and are then fine-tuned and evaluated on the same ear-EEG downstream layout across LOSO folds.
In this realistic layout-transfer regime, Device Passport improves over the strongest prior learned baseline, ACPE, for most held-out subjects (\Cref{eesm17_subject_figure}a).
The Cross-Attention and MLP variants improve over ACPE for 6--7 of 9 held-out subjects, with one additional tie, showing that the same qualitative advantage from the controlled layout-transfer experiment persists in a realistic new-device transfer setting.
The MLP expert variant is best or second-best for all 9 EESM17 subjects in the absolute-performance table (\Cref{eesm17_subject_figure}b), while the Cross-Attention variant is strongest on several subjects but less robust on the poor-contact cases.
The subjects that reportedly had poor electrode contact (Subjects 5 (our 6) and 1 (our 5) in \cite{mikkelsen2017automatic}) performed worse with Device Passport, suggesting that robustness to noisy channels is a key limitation and opportunity for improvement.
These failures are consistent with the method's dependence on the functional activity summary \(z_c\): when contact quality is poor, the adapter may receive misleading activity information even when the channel metadata are correct.
Taken together, these results suggest that metadata-guided expert embeddings can improve transfer to unseen layouts.

\section{Discussion and Conclusion}
We identify channel embeddings as an important bottleneck for reusing biosignal foundation models on unseen device layouts.
In both controlled subset-transfer experiments and full-layout transfer to ear-EEG, Device Passport is most helpful when the downstream layout differs from pretraining, suggesting that pretrained spatial information can be reused more directly than with ordinary lookup or activity-only positional encodings.
We focused on low fine-tuning sample regimes because they match the deployment setting for new biosignal devices: before large device-specific datasets exist, practitioners need pretrained models that adapt with limited labeled data.

These results support Device Passport in the layout-transfer settings we study, while leaving room for stronger robustness across all regimes.
ACPE remains a competitive baseline, and Device Passport depends on useful channel metadata.
Nevertheless, the pattern across controlled and real-device transfers suggests that metadata-guided expert embeddings provide a practical way to reuse pretrained spatial structure: downstream training can adapt the mixture rule without relearning channel representations from scratch.
The clearest remaining limitation is sensitivity to noisy or poorly contacted channels, as seen in EESM17, which points to direct extensions such as synthetic channel corruption or explicit noise experts.

Overall, Device Passport offers a lightweight mechanism for adapting pretrained biosignal models to new health-sensing devices and montages with limited device-specific data.
More broadly, these findings suggest that explicitly modeling how channel metadata and functional activity jointly define sensor identity can make biosignal foundation models more portable across real-world acquisition layouts.

\clearpage

\bibliography{ref}
\bibliographystyle{icml2026}
\newpage

\appendix
\section{Implementation details}
\label{app:implementation-details}

Unless otherwise noted, we follow the public CBraMod training defaults \cite{wang2024cbramod}. In sample-efficiency sweeps, we keep the full fine-tuning compute budget by converting the CBraMod epoch schedule into a fixed number of optimization steps for each downstream subject subset.

\begin{table}[h!]
  \centering
  \caption{Training and model settings used for the channel-embedding comparisons. Values follow the released CBraMod configuration unless overridden by the layout-transfer experiment design.}
  \label{tab:implementation-details}
  \resizebox{\columnwidth}{!}{%
    \begin{tabular}{l|l}
      Setting & Value \\
      \hline
      Backbone & CBraMod criss-cross transformer \\
      Input / embedding dimension & 200 \\
      Transformer layers / heads & 12 / 8 \\
      Feedforward dimension & 800 \\
      Temporal segments & 30 \\
      Pretraining objective & Masked patch reconstruction \\
      Pretraining mask ratio & 0.5 \\
      Pretraining epochs / batch size & 40 / 128 \\
      Pretraining optimizer schedule & AdamW with cosine LR schedule \\
      Pretraining LR / weight decay & \(5\times10^{-4}\) / \(5\times10^{-2}\) \\
      Fine-tuning epochs / batch size & 50 / 64 before step-budget matching \\
      Fine-tuning optimizer & AdamW \\
      Fine-tuning LR / weight decay & \(1\times10^{-4}\) / \(5\times10^{-2}\) \\
      Gradient clip / dropout & 1 / 0.1 \\
      Label smoothing & 0.1 \\
      Device Passport experts & 10 learned experts \\
      Downstream expert handling & Experts frozen; adapter fine-tuned \\
      Seeds & 5 pretraining seeds \(\times\) 5 fine-tuning seeds \\
      EESM17 normalization & Instance normalization \\
      \hline
    \end{tabular}%
  }
\end{table}

\section{Channel embedding details}
\label{app:channel-embeddings}

\textbf{APE.}
Absolute positional embeddings (APE) leverage sinusoidal positional encoding indexed by a channel's ordinal number in the layout \citep{vaswani2017attention}.
This can misalign spatial information when the downstream layout reorders or replaces channels.

\textbf{Channel ID.}
Learned channel embeddings are indexed by channel name (e.g., FP1) and are commonly used in neural foundation models \citep{azabou2023unified, jiang2024large}.
This works when the same 10--20 channel appears at fine-tuning time, but new devices require randomly initialized embeddings.

\textbf{XYZ.}
XYZ sinusoidal embeddings use electrode coordinates from MNE (meters) \cite{gramfort2013meg}.
We scale to millimeters, add 150~mm, round to the nearest millimeter, then index sinusoidal embeddings for each axis.
Each axis contributes 66 dimensions (200/3), concatenated and zero-padded to 200 dimensions.
This is effective with sufficient coordinate diversity \cite{chau2025population} but may be limited under single-layout pretraining.

\textbf{ACPE.}
Asymmetric conditional positional encoding (ACPE) is CBraMod's channel-aware positional encoding scheme \cite{wang2024cbramod}, adapted from conditional positional encoding (CPE) in vision \cite{chu2021conditional}.
Instead of using fixed absolute embeddings, ACPE dynamically generates positional information from the patch embeddings themselves, making it more adaptable to varying channel layouts and time lengths.
Concretely, ACPE applies a depthwise 2D convolution over the spatial (channels) and temporal dimensions of the patch-embedding grid to produce positional encodings, which are then added back to the patch embeddings.
The convolution kernel is asymmetric, using a larger spatial kernel than temporal kernel (\(k_s > k_t\)) to encode long-range spatial dependencies and shorter-range temporal dependencies, reflecting the structure of EEG signals.
This dynamic, asymmetric design has been shown to outperform APE and symmetric CPE in CBraMod when transferring across EEG formats \cite{wang2024cbramod}.

\textbf{Expert adapters (MLP / Cross-Attention).}
We learn a small set of expert embeddings during pretraining and freeze them for downstream use.
A lightweight adapter uses channel activity plus XYZ metadata to compute mixture weights over experts.
The XYZ inputs are kept in decimeter scale to match patch-embedding magnitudes.
In the MLP variant, a softmax over experts produces the channel embedding.
In the cross-attention variant, learned K/V embeddings are combined with a query projected from activity+metadata to produce the channel embedding.

\newpage

\section{ACPE vs. traditional encodings}
\label{app:acpe-vs-traditional}

\begin{figure}[h!]
  \centering
  \includegraphics[width=\linewidth]{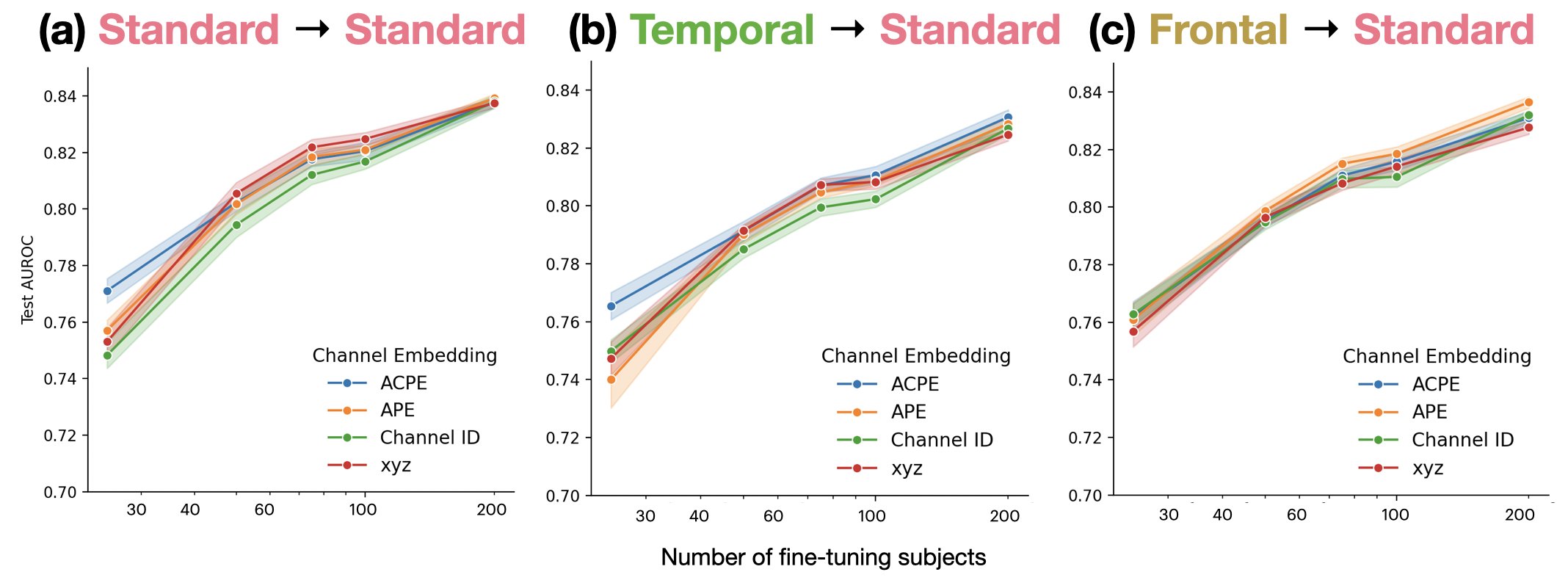}
  \caption{
  \textbf{ACPE compared with traditional positional encoding methods.}
  We compare ACPE against common channel embedding baselines (e.g., APE, Channel ID, and XYZ-based encodings) in the variable-layout pretraining setting.
  \label{acpe_vs_traditional_figure}
  }
\end{figure}

\section{EESM17 Subject Note}
\label{app:eesm17-subject-mapping}

The subject numbering used in our EESM17 plots follows our evaluation order rather than the numbering used in \cite{mikkelsen2017automatic}.
In particular, the poor-contact subjects called out in the main text correspond to Subject 5 in \cite{mikkelsen2017automatic} (our Subject 6) and Subject 1 in \cite{mikkelsen2017automatic} (our Subject 5).

\end{document}